\documentclass[conference]{IEEEtran}  
\pdfoutput=1 
\IEEEoverridecommandlockouts
\usepackage{amsmath,amssymb,amsfonts}
\usepackage{algorithmic}
\usepackage{graphicx}
\usepackage{textcomp}
\usepackage{xcolor}
\usepackage{lipsum}
\usepackage{graphicx}
\graphicspath{{figures/}}
\usepackage{booktabs}
\usepackage{color}
\usepackage{amsmath}
\usepackage{amssymb}  
\usepackage{booktabs} 
\usepackage{multirow} 
\usepackage{tabularx} 
\usepackage{rotating}
\usepackage{caption}
\def\BibTeX{{\rm B\kern-.05em{\sc i\kern-.025em b}\kern-.08em
    T\kern-.1667em\lower.7ex\hbox{E}\kern-.125emX}}
    
\begin{document}

\title{Efficient Online Learning for Networks of Two-Compartment Spiking Neurons\\

}

\author{\IEEEauthorblockN{Yujia Yin}
\IEEEauthorblockA{\textit{Department of Computing} \\
\textit{The Hong Kong Polytechnic University}\\
Hong Kong SAR, China \\
yujia.yin@connect.polyu.hk}
\and
\IEEEauthorblockN{Xinyi Chen}
\IEEEauthorblockA{\textit{Department of Computing} \\
\textit{The Hong Kong Polytechnic University}\\
Hong Kong SAR, China\\
xinyi-97.chen@connect.polyu.hk}
\and
\IEEEauthorblockN{Chenxiang Ma}
\IEEEauthorblockA{\textit{Department of Computing} \\
\textit{The Hong Kong Polytechnic University}\\
Hong Kong SAR, China\\
chenxiang.ma@connect.polyu.hk}
\and
\IEEEauthorblockN{Jibin Wu*}
\IEEEauthorblockA{\textit{Department of Computing} \\
\textit{The Hong Kong Polytechnic University}\\
Hong Kong SAR, China\\
jibin.wu@polyu.edu.hk}
\thanks{*Corresponding Author}
\and
\IEEEauthorblockN{Kay Chen Tan}
\IEEEauthorblockA{\textit{Department of Computing} \\
\textit{The Hong Kong Polytechnic University}\\
Hong Kong SAR, China\\
kctan@polyu.edu.hk}
\thanks{This manuscript is under review by 2024 International Joint Conference on Neural Networks (IJCNN 2024).}
}


\maketitle

\begin{abstract}

The brain-inspired Spiking Neural Networks (SNNs) have garnered considerable research interest due to their superior performance and energy efficiency in processing temporal signals. Recently, a novel multi-compartment spiking neuron model, namely the Two-Compartment LIF (TC-LIF) model, has been proposed and exhibited a remarkable capacity for sequential modelling. However, training the TC-LIF model presents challenges stemming from the large memory consumption and the issue of gradient vanishing associated with the Backpropagation Through Time (BPTT) algorithm. To address these challenges, online learning methodologies emerge as a promising solution. Yet, to date, the application of online learning methods in SNNs has been predominantly confined to simplified Leaky Integrate-and-Fire (LIF) neuron models. In this paper, we present a novel online learning method specifically tailored for networks of TC-LIF neurons. Additionally, we propose a refined TC-LIF neuron model called Adaptive TC-LIF, which is carefully designed to enhance temporal information integration in online learning scenarios. Extensive experiments, conducted on various sequential benchmarks, demonstrate that our approach successfully preserves the superior sequential modeling capabilities of the TC-LIF neuron while incorporating the training efficiency and hardware friendliness of online learning. As a result, it offers a multitude of opportunities to leverage neuromorphic solutions for processing temporal signals.

\end{abstract}

\section{Introduction}
The human brain, with a power consumption of only 20 Watts, is capable of processing complex sensory signals efficiently \cite{markram2006blue}. Mimicking the discrete, spike-based information processing observed in the brain, spiking neural networks (SNNs) have been introduced for artificial intelligence (AI) applications \cite{yang2022deep,wu2018spiking,wu2021tandem}.  Remarkably, owing to the introduction of the backpropagation through time (BPTT) learning algorithm with surrogate gradients \cite{neftci2019surrogate, stbp}, 
SNNs have achieved promising performance across a wide range of AI tasks \cite{wu2021progressive, meng2022training, hybridcoding}. Moreover, these models demonstrate superior energy efficiency when deployed on ultra-low-power neuromorphic hardware \cite{pei2019towards, davies2018loihi, merolla2014million}.

However, current SNNs still face challenges in effectively processing sensory signals compared to their biological counterparts \cite{gutig2016spiking, chen2023unleashing}. One contributing factor is the use of overly simplified neuron models, such as the widely adopted Leaky Integrate-and-Fire (LIF) model~\cite{maass1997networks,lif}, which has limited capacity to store information over long time windows. To overcome this limitation, recent research has introduced slow-decaying variables into spiking neuron models to enrich their neuronal dynamics and improve their ability to process complex temporal signals \cite{yin2021accurate, rao2022long, DEXAT}. Yet, these models struggle to establish long-term temporal dependencies that are crucial for effective temporal signal processing. Another line of research has explored the use of self-attention mechanisms to enable more flexible integration of temporal information within SNNs \cite{TASNN,TASNN2, qin2023attention}. However, these models require significant memory resources that are currently unavailable on neuromorphic chips. Recently, a two-compartment LIF (TC-LIF) neuron model has been proposed, which captures the intricate neuronal structure and rich temporal dynamics observed in biological neurons \cite{zhang2023tc}. By incorporating the interaction between the somatic and dendritic compartments, the TC-LIF neuron demonstrates superior sequential modeling capabilities in many challenging temporal processing tasks.

\begin{figure*}[ht]
    \centering
    \includegraphics[width = 1\linewidth, trim=0 70 0 20mm, clip]{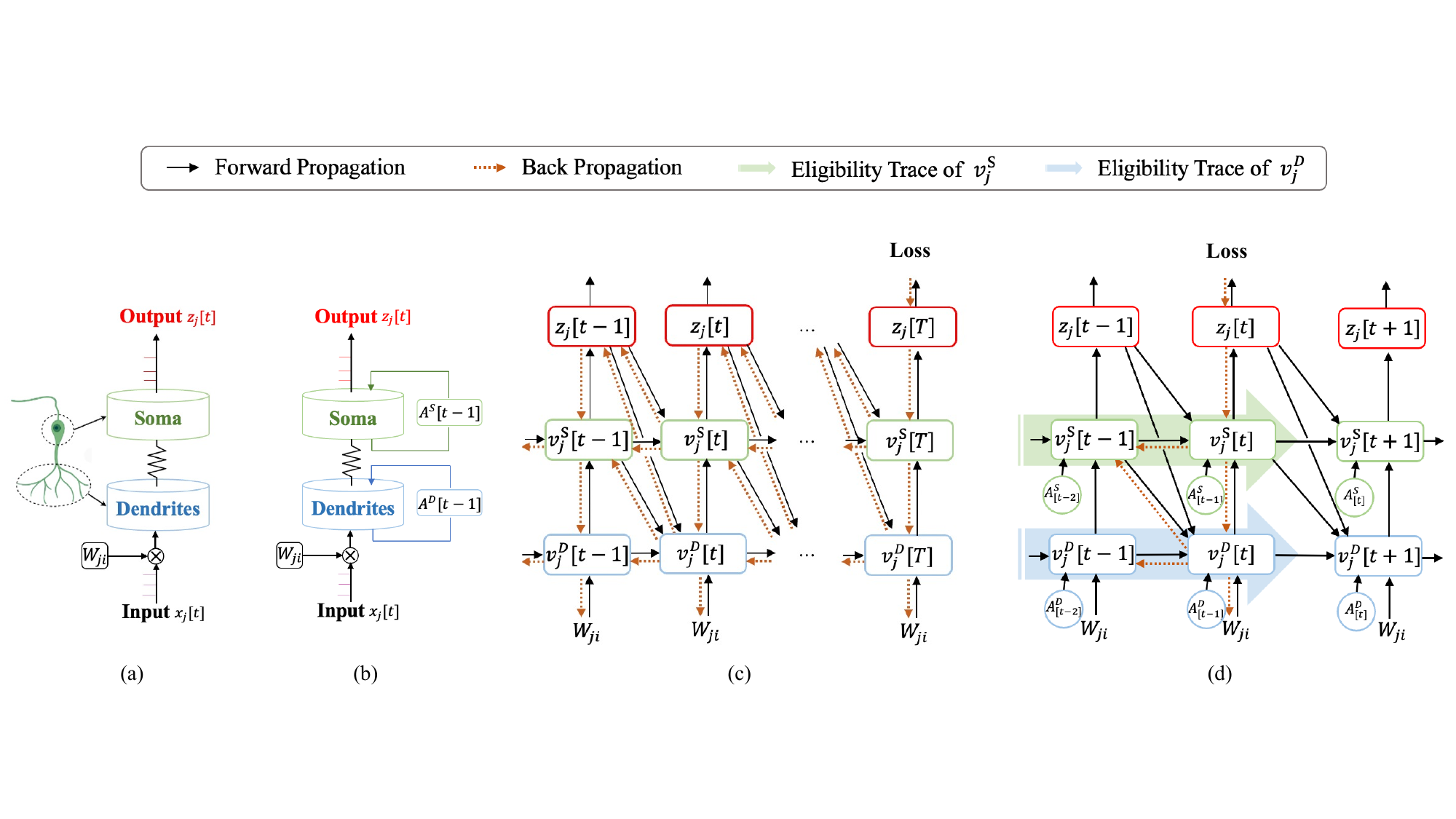}
    
    \caption{Comparison of the proposed neuron models and learning algorithms. \textbf{(a)} The vanilla TC-LIF neuron captures the interaction between dendritic and somatic compartments in biological neurons to facilitate temporal signal processing. \textbf{(b)} The proposed Adaptive TC-LIF neuron further introduces time-varying membrane potentials decaying constants to facilitate temporal information integration during online learning. \textbf{(c)} The vanilla TC-LIF model is trained using BPTT algorithm, where the gradients are propagated from the last time step to all preceding time steps for parameter update. \textbf{(d)} Our proposed Adaptive TC-LIF neuron facilitates efficient online learning by computing parameter updates at each time step based on the local loss and eligibility trace derived at that specific time step. Importantly, the computation of eligibility traces for both neuronal compartments occurs in a forward manner, eliminating the need to store intermediate network states as required in BPTT training.}
    \label{1}
\end{figure*}

Despite the impressive performance of SNNs utilizing TC-LIF neurons in sequential modeling, they face significant challenges during BPTT training. As depicted in Fig. \ref{1}, the BPTT algorithm necessitates storing all internal states over the whole time window for gradient updates, leading to a linear increase in memory requirements as time window increases. Additionally, the BPTT algorithm is susceptible to issues like vanishing or exploding gradients \cite{zheng2021going,wang2023adaptive}, which further complicates the task of stable training due to the necessity of carefully initializing the parameters. Furthermore, the BPTT algorithm requires the entire sequence to be available before training, making it incompatible with online learning scenarios {\cite{xiao2022online}, \cite{bellec2020solution}, \cite{9328869}}. 

To tackle these challenges, this study explores the biologically plausible and efficient online training approach known as e-prop \cite{bellec2020solution}. The e-prop algorithm is based on the concept of eligibility trace, which enables the local and feedforward computation of the error gradient at each time step. Notably, e-prop has demonstrated comparable performance to BPTT on a number of AI benchmarks \cite{manneschi2020alternative,frenkel2022reckon}. However, it is worth noting that this method has primarily been investigated using simplified LIF neurons, and its applicability to more complex multi-compartment neuron models has yet to be explored {\cite{yin2021accurate}}. Notably, the TC-LIF neuron was initially designed to work with the BPTT training algorithm, which necessitates the adaptation of its parameter space when transitioning to online learning settings. 

In this study, we present an extension of the e-prop formulation to accommodate the unique neuronal dynamics of the TC-LIF neuron model. Additionally, we propose an Adaptive TC-LIF model specifically designed for online learning scenarios. The traditional BPTT algorithm suffers from the issue of gradient vanishing, which restricts the leakiness of the membrane potential in the original TC-LIF model. However, the introduction of the e-prop algorithm addresses this limitation, allowing for more flexible selection of the membrane potential decaying constants. To ensure a balanced influence of inputs from different time steps on parameter updates, we further propose to set the membrane potential decaying constants to be time-varying. Experimental results demonstrate that our proposed approach successfully preserves the superior sequential modeling ability of TC-LIF neurons while incorporating the efficiency of online learning. The main contributions of this work can be summarized as follows:
\begin{itemize}
    \item We expand upon the e-prop online learning approach, which was initially designed for LIF neurons, and adapt it to accommodate multi-compartment TC-LIF neurons. We provide comprehensive and rigorous mathematical derivations for the proposed online learning method.

    \item We propose a novel Adaptive TC-LIF model that incorporates time-varying membrane potential decaying constants. This enhances temporal information integration in online learning scenarios.

    \item We perform comprehensive experiments on a range of sequential modeling tasks to evaluate the performance of our proposed Adaptive TC-LIF model. The results substantiate that our model exhibits exceptional sequential modeling capacity, high training efficiency, and neuromorphic hardware friendliness.
    
\end{itemize}

\section{Preliminaries}
\subsection{Spiking Neuron Models}
Motivated by the intricate structure and dynamic firing behavior of biological neurons, numerous spiking neuron models have been proposed in the literature \cite{gerstner2014neuronal, fang2021incorporating, GLIF}. These models offer SNNs superior capabilities in processing temporal signals. Typically, the neuronal dynamics of these spiking neurons encompass three fundamental components: charging, firing, and resetting. For instance, the discrete-time equations below can be used to model recurrently connected \cite{yin2021accurate, yin2020effective} LIF neurons:
\begin{align}
v_j[t+1] &= \alpha v_j[t] - v_{th} z_j[t] + I_j[t+1], \\
I_j[t+1] &= \sum_{{i\neq j}} W_{ji}^{rec} z_i[t] + \sum_{{i}} W_{ji}^{in} x_i[t+1] + b,\quad \\
z_j[t+1] &= H(v_j[t+1] - v_{th}), \quad 
\end{align}
where $W_{ji}^{in}$ and $W_{ji}^{rec}$ denote the feedforward and recurrent connections from the presynaptic neuron $i$ to the postsynaptic neuron $j$, respectively. The parameter $\alpha$ represents the membrane decaying coefficient, which ranges between 0 and 1. The variable $I_j[t+1]$ represents the input current at time $t+1$. During the charging process, the membrane potential $v_j[t]$ is updated until it reaches the firing threshold $v_{th}$, at which point an output spike $z_j[t]$ is generated. Subsequently, after firing, the membrane potential is reset.


While the LIF neuron demonstrates promising results in tasks with limited temporal context, it encounters difficulties in retaining information over longer time periods (e.g., a few hundred time steps), resulting in poor performance for tasks that require long-term memory. This limitation arises from the exponential decay of inputs to LIF neurons during their membrane potential update. To enhance the sequential modeling capacity of SNNs, a recent study has introduced a novel spiking neuron model called TC-LIF \cite{zhang2023tc}. The TC-LIF neuron is specifically designed to replicate the interaction between dendritic and somatic compartments of biological neurons. The neuronal dynamics of the TC-LIF neuron can be formulated as follows:
\begin{align} \label{eq:update_rule_vD}
v_j^D[t+1] &= v_j^D[t] + \beta_1 v_j^S[t] - \gamma z_j[t] + I_j[t+1], \\
\label{eq:update_rule_vS}
v_j^S[t+1] &= v_j^S[t] + \beta_2 v_j^D[t+1] - v_{th} z_j[t], \\
\label{eq:update_rule_I}
I_j[t+1] &= \sum_{{i\neq j}} W_{ji}^{rec} z_i[t] + \sum_{{i}} W_{ji}^{in} x_i[t+1] + b, \\
\label{eq:activation_function}
z_j[t+1] &= H(v_j^S[t+1] - v_{th}).
\end{align}

The TC-LIF neuron exhibits two distinct neuronal states, denoted as $h_j[t] = [v_j^D[t], v_j^S[t]]$. Notably, $v_j^D[t]$ corresponds to the dendritic compartment, which is responsible for storing long-term memory. On the other hand, $v_j^S[t]$ represents the somatic compartment, which is used to store short-term memory. The interaction between these two compartments is regulated by the coefficients $\beta_1$ and $\beta_2$. Specifically, $\beta_1$ is defined as $-\sigma(c_1)$, and $\beta_2$ is defined as $\sigma(c_2)$, where $\sigma(\cdot)$ denotes a sigmoid function. Both $c_1$ and $c_2$ are learnable parameters that are jointly trained with network weights. Following the somatic firing, both compartments undergo a reset in their membrane potential. It is worth noting that the reset of the dendritic compartment involves the initiation of a backpropagating spike, the effect of which is governed by the hyperparameter $\gamma$.


\subsection{Challenges of Training SNNs with BPTT}
Training SNNs presents a significant challenge due to the intricate spatial-temporal dependency between distant spikes and the discrete nature of spike generation. However, recent advancements have addressed these issues by applying BPTT algorithm~\cite{werbos1990backpropagation} coupled with the surrogate gradient method. To illustrate how BPTT algorithm can be applied across LIF and TC-LIF spiking neurons, we denote the observable state (i.e., spike) of neuron $j$ at time $t$ as $z_j[t]$, the hidden state as $h_j[t]$ (i.e., membrane potential), and the input as $x_j[t]$. The update of the hidden state of a neuron can be expressed as $h_j[t] = f(h_j[t-1], z_j[t-1], x_j[t], W_{ji})$. The weight gradients can be expressed as follows:
\begin{equation}
\label{eq:BPTT}
\frac{dE}{dW_{ji}} = \sum_{t=1}^{T} \frac{dE}{dz_j[T]} \frac{\partial z_j[T]}{\partial W_{ji}},   
\end{equation}
where $E$ represents the error term. It is worth noting that by expanding the above gradient expression using the chain rule, as shown in (\ref{eq:BPTT_expand}), it becomes apparent that the BPTT algorithm optimizes SNN parameters by unfolding the network state over time and then applying the backpropagation algorithm. However, this process requires the storage of intermediate states for all neurons, which imposes a significant computational burden. 
\begin{equation}
\label{eq:BPTT_expand}
\frac{dE}{dW_{ji}} = \sum_{t=1}^T \frac{dE}{d z_{j}[T]} \frac{\partial z_{j}[T]}{\partial h_{j}[T]} \frac{\partial h_{j}[T]}{\partial h_{j}[t]} \frac{\partial h_{j}[t]}{\partial W_{ji}}.
\end{equation}


\subsection{Online learning with e-prop algorithm}

The e-prop algorithm \cite{bellec2020solution} has emerged as an alternative to the BPTT training algorithm. This biologically inspired online learning algorithm consists of two essential components: learning signals and eligibility traces. The learning signals encompass various top-down signals in the brain, such as neuromodulators and error-related neural activities. These signals convey information to neuronal populations about behavioral outcomes. On the other hand, eligibility traces refer to molecular-level traces exist within each neuron, such as calcium ions or activated CaMKII enzyme. These traces represent fading memories of past events. Supported by rigorous mathematical derivation~\cite{bellec2020solution}, the effective integration of eligibility trace and learning signals can lead to high-performance online learning.

\begin{figure}[ht]
    \centering
    \includegraphics[width = \linewidth, trim=40 0 40 0mm, clip]{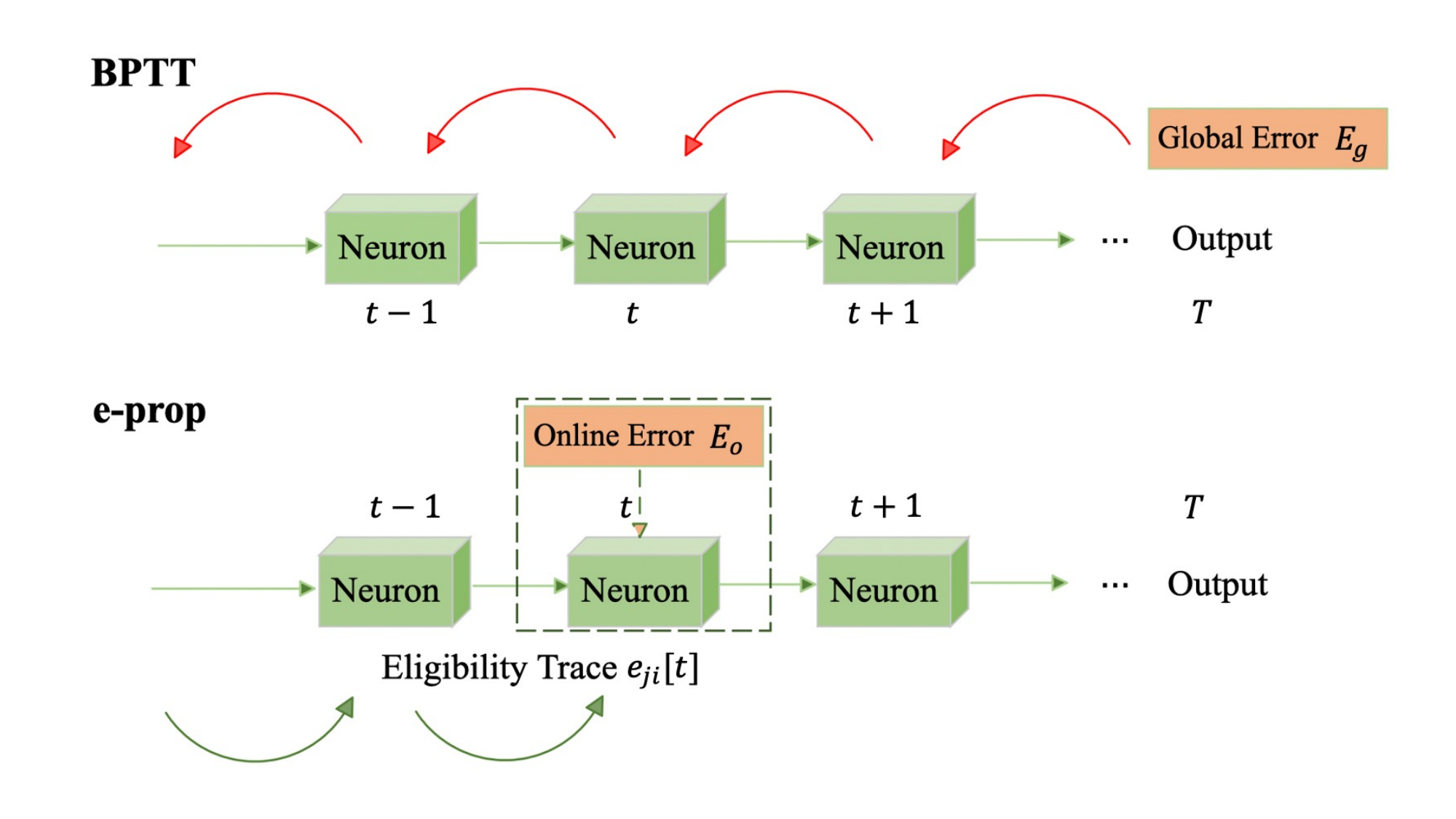}
    \caption{The comparison of gradient update principle of BPTT and e-prop algorithms at time $t$. BPTT algorithm updates the network parameters based on the global loss, which can only be obtained at the last time step $T$. In contrast, the e-prop algorithm accumulates eligibility traces that propagate forward in time and multiplies them with the online error term. This allows for the online update of network parameters.}
    \label{compare BPTT&e-prop}
\end{figure}

As given in (\ref{eq:local}), the impact of future time steps from ${t+1}$ to $T$ on the present state $t$ is disregarded in e-prop, allowing the weight gradient $\frac{dE}{dW_{ji}}$ to be expressed as a summation of products across time steps $1$ to $t$. The second term of this equation represents a local gradient that remains independent of error $E$:

\begin{equation}
\label{eq:local}
\frac{dE}{dW_{ji}} = \sum_{t} \frac{dE}{dz_j[t]} \left(\frac{d z_j[t]}{d W_{ji}}\right)_{local}.
\end{equation}

In e-prop, the component $\frac{dE}{dz_j[t]}$ is further approximated with an online learning signal $L_j[t] \triangleq \frac{\partial E}{\partial z_j[t]}$, where $E$ is the cross entropy loss. Furthermore, the eligibility traces $e_{ji}[t] \triangleq \left(\frac{d z_j[t]}{d W_{ji}}\right)_{local}$ aim to capture the maximum amount of information about the network gradient, which can be updated during the forward computation. Consequently, the computation of the gradient in e-prop involves a combination of the learning signal and the eligibility trace as:
\begin{align}
\label{weight gradient}
\frac{dE}{dW_{ji}} = \sum_{t} L_j[t] \cdot e_{ji}[t]. 
\end{align}

In the case where the learning signal assumes an ideal value, we can provide rigorous proof demonstrating that the eligibility trace effectively preserves the maximum amount of gradient information \cite{bellec2020solution}. In the e-prop algorithm, the derivation of the eligibility trace accounts for the evolution process of the hidden state within the neuron, eliminating the need to store all hidden states by disregarding the influence of future time steps. The specific expression of the eligibility trace is:
\begin{equation}
\label{eq:etrace}
e_{ji}[t]
= \frac{\partial z_{j}[t]}{\partial h_{j}[t]} \substack{\underbrace{\sum_{t^\prime \leq t}  \frac{\partial h_{j}[t]}{\partial h_{j}[t -1]} \cdots  \frac{\partial h_{j}[t^\prime+1]}{\partial h_{j}[t^\prime]} \frac{\partial h_{j}[t^\prime]}{\partial W_{ji}}} \\ \vphantom{\sum_{t^\prime \leq t}} = \varepsilon_{ji}[t]}.
\end{equation}
In \eqref{eq:etrace}, the first term represents the surrogate gradient $\psi_j[t]$ \cite{neftci2019surrogate}, and the following terms are defined as eligibility vector $\varepsilon_{ji}[t]$. Furthermore, we can derive an iterative expression for the eligibility vector that pertains to the eligibility trace $e_{ji}[t] = \psi_j[t] \cdot \varepsilon_{ji}[t]$ as:

\begin{equation}
\begin{aligned}
\label{eligibility vector}
\varepsilon_{ji}[t] &\triangleq \sum_{t^\prime \leq t}  \frac{\partial h_{j}[t]}{\partial h_{j}[t -1]} \cdots  \frac{\partial h_{j}[t^\prime+1]}{\partial h_{j}[t^\prime]} \frac{\partial h_{j}[t^\prime]}{\partial W_{ji}} 
\\
&= \frac{\partial h_{j}[t]}{\partial W_{ji}} + \frac{\partial h_{j}[t]}{\partial h_{j}[t-1]} \frac{\partial h_{j}[t-1]}{\partial W_{ji}} + \cdots 
\\
&+ \frac{\partial h_{j}[t]}{\partial h_{j}[t-1]} \cdots \frac{\partial h_{j}[2]}{\partial h_{j}[1]} \frac{\partial h_{j}[1]}{\partial W_{ji}}
\\
&= \frac{\partial h_{j}[t]}{\partial W_{ji}} + \frac{\partial h_{j}[t]}{\partial h_{j}[t-1]} \cdot
\\
&\left(\frac{\partial h_{j}[t-1]}{\partial W_{ji}} 
+ \frac{\partial h_{j}[t-1]}{\partial h_{j}[t-2]} \frac{\partial h_{j}[t-2]}{\partial W_{ji}} +\cdots \right)
\\
&= \frac{\partial h_{j}[t]}{\partial h_{j}[t-1]} \varepsilon_{ji}[t -1] + \frac{\partial h_{j}[t]}{\partial W_{ji}}.
\end{aligned}
\end{equation}

It is evident that the key aspect of the e-prop algorithm revolves around the derivation of the neuron's eligibility vector, which holds a substantial amount of gradient information and is propagated forward in time.

\section{Methodology}

\subsection{Eligibility Traces for TC-LIF}

Here, we present an elaborate derivation for the eligibility vector of the TC-LIF neuron. By combining the neuronal function of the TC-LIF neuron and the eligibility trace defined in the earlier section, we can obtain the eligibility trace of the dendritic compartment as:
\begin{equation}
\scalebox{1}{
$\begin{aligned}
\frac{\partial v_j^D[t]}{\partial W_{ji}^{in}} 
&= \frac{\partial v_j^D[t]}{\partial W_{ji}^{in}} + \frac{\partial v_j^D[t]}{\partial v_j^D[t-1]} \frac{\partial v_j^D[t-1]}{\partial W_{ji}^{in}} 
\\
&+ \frac{\partial v_j^D[t]}{\partial v_j^S[t-1]} \frac{\partial v_j^S[t-1]}{\partial W_{ji}^{in}}
\\
&= x_i[t] + \frac{\partial v_j^D[t-1]}{\partial W_{ji}^{in}} + \beta_1 \frac{\partial v_j^S[t-1]}{\partial W_{ji}^{in}}.
\end{aligned}
$}
\label{dddw}
\end{equation}

Similarly, for the somatic compartment, the eligibility trace is formulated as follows:
\begin{equation}
\scalebox{1}{
$\begin{aligned}
\frac{\partial v_j^S[t]}{\partial W_{ji}^{in}} 
&= \frac{\partial v_j^S[t]}{\partial v_j^S[t-1]} \frac{\partial v_j^S[t-1]}{\partial W_{ji}^{in}} + \frac{\partial v_j^S[t]}{\partial v_j^D[t]} \frac{\partial v_j^D[t]}{\partial W_{ji}^{in}}
\\
&= \frac{\partial v_j^S[t-1]}{\partial W_{ji}^{in}} + \beta_2 \frac{\partial v_j^D[t]}{\partial W_{ji}^{in}}.
\end{aligned}
$}
\label{dsdw}
\end{equation}

The Eqs. (\ref{dddw}) and (\ref{dsdw}) can be further abstracted into a two-dimensional eligibility vector $\varepsilon_{ji}[t] = [\varepsilon_{ji}^{D}[t],\varepsilon_{ji}^{S}[t]]=[\frac{\partial v_j^D[t]}{\partial W_{ji}^{in}},\frac{\partial v_j^S[t]}{\partial W_{ji}^{in}}]$ and expressed iteratively as follows:
\begin{align}
\label{ed}
\varepsilon_{ji}^{D}[t] &= \varepsilon_{ji}^{D}[t-1] + \beta_1 \varepsilon_{ji}^{S}[t-1] + x_i[t],
\\
\label{es}
\varepsilon_{ji}^{S}[t] &= \varepsilon_{ji}^{S}[t-1] + \beta_2 \varepsilon_{ji}^{D}[t].
\end{align}

By plugging \eqref{ed} into \eqref{es}, we could rewrite the eligibility vector as:
\begin{align}
\label{eq:signalTC1}
\varepsilon_{ji}^{D}[t] &= \varepsilon_{ji}^{D}[t-1] + \beta_1 \varepsilon_{ji}^{S}[t-1] + x_i[t],\\
\label{eq:signalTC2}
\varepsilon_{ji}^S[t] &= (1+ \beta_1 \beta_2) \varepsilon_{ji}^{S}[t-1] + \beta_2 \varepsilon_{ji}^{D}[t-1] + \beta_2 x_i[t]. 
\end{align}

As the eligibility trace defined in (\ref{eq:etrace}), we are left with the task of deriving $\frac{\partial z_{j}[t]}{\partial h_{j}[t]}$, which can be obtained as follows: 
\begin{equation}
\begin{aligned}
\frac{\partial z_{j}[t]}{\partial h_{j}[t]} &\triangleq [\frac{\partial z_{j}[t]}{\partial v_{j}^{D}[t]},\frac{\partial z_{j}[t]}{\partial v_{j}^{S}[t]}] = \left[ \frac{\partial z_{j}[t]}{\partial v_{j}^{S}[t]}\frac{\partial v_{j}^{S}[t]}{\partial v_{j}^{D}[t]}, \frac{\partial z_{j}[t]}{\partial v_{j}^{S}[t]}\right] 
\\
&= [\beta_2 \psi_j[t], \psi_j[t]],    
\end{aligned}   
\end{equation}
with 
\begin{align}
\psi_j[t] = \frac{\partial z_{j}[t]}{\partial v_{j}^{S}[t]} = \frac{1}{\gamma^2} max(0,\gamma -\lvert v^{S}_j[t]-v_{th} \rvert),
\end{align}
which is the surrogate gradient~\cite{deng2022temporal} function used to overcome the issue of discontinuity arising from spike generation and reset.

As a result, the eligibility trace of the TC-LIF neuron can be calculated as:
\begin{equation}
\begin{aligned}
e_{ji}[t] 
&= \frac{\partial z_{j}[t]}{\partial h_{j}[t]} \cdot \varepsilon_{ji}[t]
\\
&= \psi_j[t](\beta_2 \varepsilon_{ji}^{D}[t] + \varepsilon_{ji}^{S}[t]).
\end{aligned}   
\end{equation}

\subsection{Redesign TC-LIF for Effective Online Learning}
\subsubsection{Redesigned Parameter Space}
The original TC-LIF neuron model has been meticulously designed to work with the BPTT algorithm, wherein the membrane decaying constants of the two neuronal compartments, namely $\alpha_1$ and $\alpha_2$, are set to $1$ to ensure the preservation of gradient information across a long time window. Such a design is critical for circumventing the vanishing gradient problem commonly encountered in BPTT training. However, within the context of e-prop online learning, the gradient preservation strategy is unnecessary. Unlike BPTT, which relies on the chain rule for gradient updates, e-prop employs an eligibility trace that updates iteratively for online parameter update. As a result, e-prop algorithm is not susceptible to the catastrophic vanishing gradient problem. Hence, we can relax the parameter space of the vanilla TC-LIF neuron to allow the values of $\alpha_1$ and $\alpha_2$ to become learnable. This modification can significantly enhance the representation power of TC-LIF neurons. 
As a result, the modified TC-LIF neuron can be expressed as:
\begin{align}
v_j^D[t+1] &= \alpha_1 v_j^D[t] + \beta_1 v_j^S[t] - \gamma z_j[t] + I_j[t+1],\quad  \\
v_j^S[t+1]  &= \alpha_2 v_j^S[t] + \beta_2 v_j^D[t+1] - v_{th} z_j[t], \quad  \\
I_j[t+1] &= \sum_{{i\neq j}} W_{ji}^{rec} z_i[t] + \sum_{{i}} W_{ji}^{in} x_i[t+1] +b, \quad  \\
z_j[t+1] &= H(v_j^S[t+1] - v_{th}), \quad 
\end{align}
where the value of $\beta_1$ and $\beta_2$ are fixed.

Furthermore, this redesigned parameter space of $\alpha_1$ and $\alpha_2$ can also facilitate the memory integration process by balancing the gradient contributions from different time steps. This point is clearer if we look at the eligibility trace of the somatic compartment of the original TC-LIF neuron as expressed in \eqref{eq:signalTC2}. It is obvious that equal importance $\beta_2$ has been assigned to both the input term $x_i[t]$ and the memory term $\varepsilon_{ji}^{D}[t-1]$. However, we argue that this is not a reasonable strategy. Specifically, the memory term $\varepsilon_{ji}^{D}[t-1]$ defined in \eqref{eq:signalTC1} consistently integrates the inputs $x_i[i]$ from $i=0$ to $t-1$ without any decay, and its impact on weight parameters in \eqref{weight gradient} will also accumulate over time. As a result, the inputs from earlier time steps' will have a much greater impact than the later time steps, and the model will gradually lose sensitivity to new input data as time progresses. Hence, it is imperative to re-weight the impact of $\varepsilon_{ji}^{D}[t-1], \varepsilon_{ji}^{S}[t-1]$, and $x_i[t]$ in \eqref{eq:signalTC2} to ensure the inputs at different time step can have a more balanced contribution. This issue can be effectively alleviated in our redesigned parameter space described as follows:
\begin{align}
\varepsilon_{ji}^{D}[t] &= \alpha_1 \varepsilon_{ji}^{D}[t-1] + \beta_1 \varepsilon_{ji}^{S}[t-1] + x_i[t],
\\
\label{eq:adapt_tc2}
\varepsilon_{ji}^S[t] &= (\alpha_2 + \beta_1 \beta_2) \varepsilon_{ji}^{S}[t-1] + \alpha_1\beta_2 \varepsilon_{ji}^{D}[t-1] + \beta_2 x_i[t],    
\end{align}
where the importance of each item can be independently adjusted by adapting the value of the membrane potential decaying constants $\alpha_1, \alpha_2$.  

\subsubsection{Time-varying Decaying Constant}

\begin{figure*}[t]
\centering
\begin{minipage}[t]{0.34\linewidth}
\raggedleft
\includegraphics[scale= 1,height=5cm]{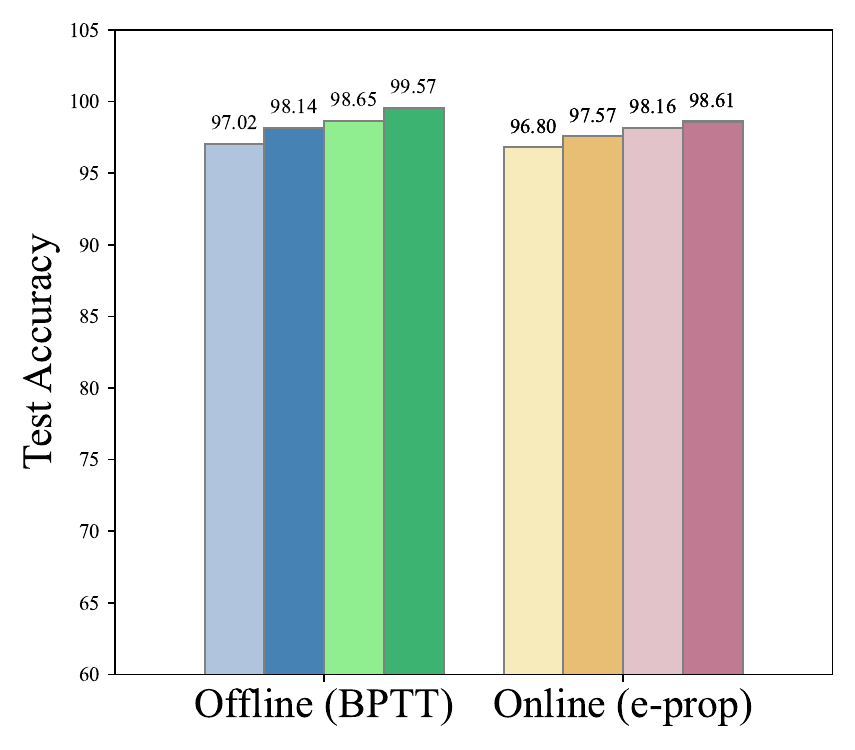}
\caption*{(a) S-MNIST} 
\end{minipage}\hfill
\begin{minipage}[t]{0.33\linewidth}
\centering
\includegraphics[scale= 1,height=5cm]{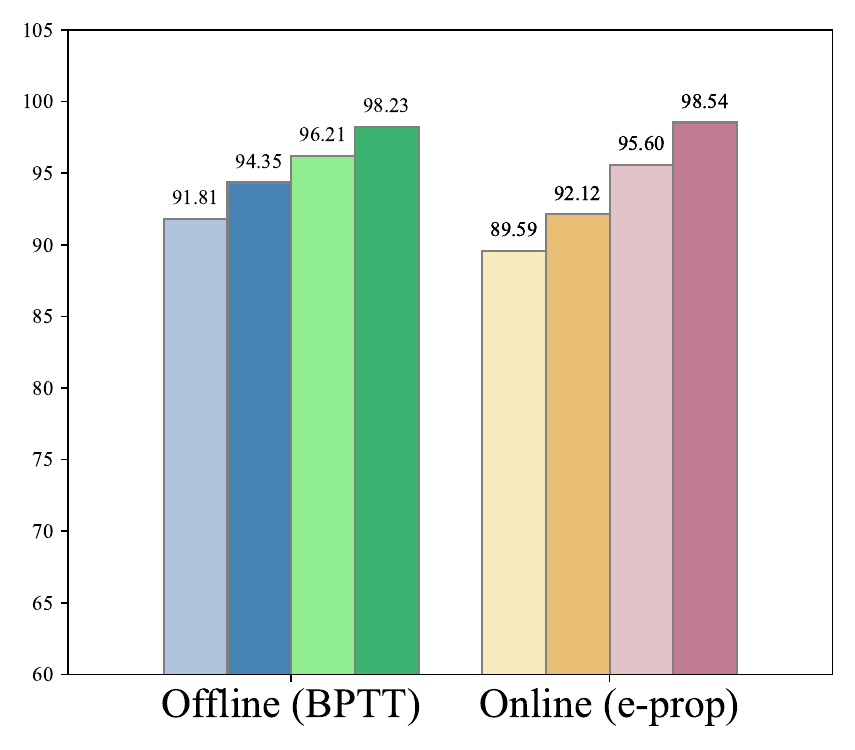}
\caption*{(b) PS-MNIST} 
\end{minipage}\hfill
\begin{minipage}[t]{0.33\linewidth}
\raggedright
\includegraphics[scale= 1,height=5cm]{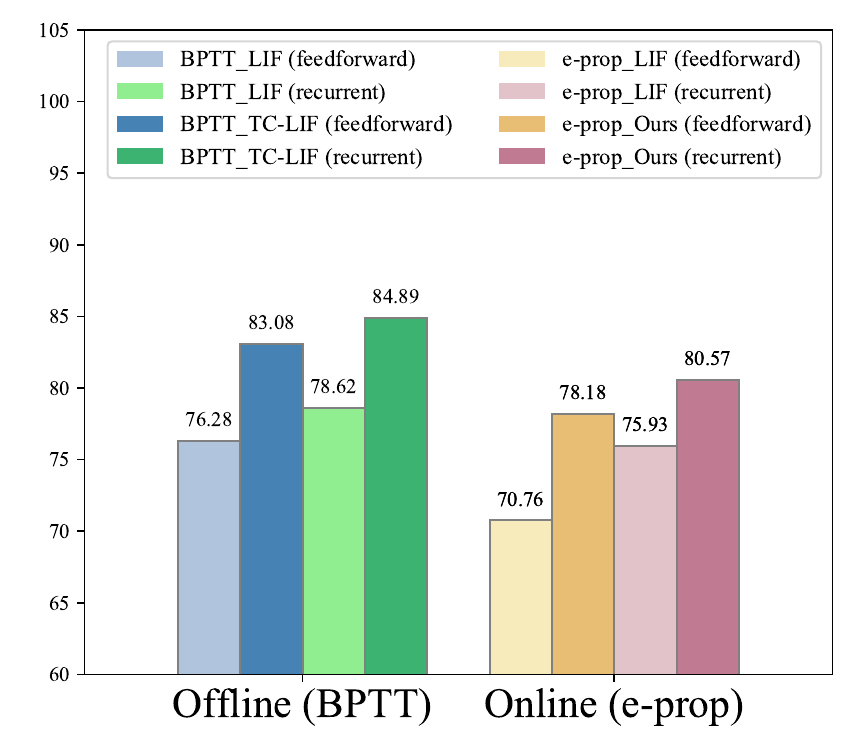}
\caption*{(c) SHD}
\end{minipage}
\caption{Comparison of model performance on three benchmark tasks.}
\label{Fig:superior classification capability}
\end{figure*}

As previously discussed, it is essential to balance the influence of inputs from different time steps during the gradient calculation. To further improve the sensitivity on new inputs, we should gradually reduce the contribution of $\varepsilon_{ji}^{D}[t-1], \varepsilon_{ji}^{S}[t-1]$ in \eqref{eq:adapt_tc2} as the training progresses. In order to achieve this, we introduce modifications to the decaying constants $\alpha_1$ and $\alpha_2$ of the membrane potentials, making them time-varying. Specifically, we use two time-varying terms $A^D[t], A^S[t]$ to describe the memory trace decaying rates for the dendritic and somatic compartments, respectively, and we call the resulted neuron model as \textbf{Adaptive TC-LIF model}. The neuronal dynamics of the Adaptive TC-LIF model can be sumamrised as:
\begin{align}
v_j^D[t+1] &= A^D[t] \cdot v_j^D[t] + \beta_1 v_j^S[t] - \gamma z_j[t] + I_j[t+1],\quad  \\
v_j^S[t+1]  &= A^S[t] \cdot v_j^S[t] + \beta_2 v_j^D[t+1] - v_{th} z_j[t], \\
A^D[t] &=Clamp\ (\Gamma(t+1,1/(t+1)),\ a^d,\ 1),\\
A^S[t] &=Clamp\ (\Gamma(t+1,1/(t+1)),\ a^s,\ 1),
\end{align}
where $\Gamma(\cdot)$ is the gamma distribution function, which follows the long-tail distribution of memory trace discovered in cerebral cortical neurons \cite{bernacchia2011reservoir}. 
$a^d$ and $a^s$ are two learnable coefficients, describing the minimum value of $A^D[t], A^S[t]$, respectively. 




\subsection{Extending e-prop to Multi-layer SNNs}
The original e-prop algorithm is proposed for single-layer networks. In this work, we extend this algorithm to multi-layer networks to solve more complex sequential tasks \cite{ma2022deep}. With a single-layer network architecture, the original weight update rule of e-prop can be formulated as:

\begin{equation}
\label{outputlayer}
\Delta W_{ji}^{o}[t]= -\eta L_j[t] \frac{\partial z_{j}^{o}[t]}{\partial h_{j}^{o}[t]} \frac{\partial h_{j}^{o}[t]}{\partial W_{ji}^{o}}.
\end{equation}
where $\eta$ denotes the learning rate, the superscript ``$o$'' indicates the output layer, and $\frac{\partial h_{j}^{o}[t]}{\partial W_{ji}^{o}}$ represents the eligibility vector of the output layer. We can straightforwardly extend (\ref{outputlayer}) to the hidden layers of multi-layer networks as:
\begin{equation}
\Delta W_{ji}^{l}[t]
= -\eta L_j[t] \frac{\partial z_{j}^{o}[t]}{\partial h_{j}^{o}[t]} \frac{\partial h_{j}^{o}[t]}{\partial z_{j}^{l}[t]} \frac{\partial z_{j}^{l}[t]}{\partial h_{j}^{l}[t]} \frac{\partial h_{j}^{l}[t]}{\partial W_{ji}^{l}},
\end{equation}
where the superscript ``$l$'' denotes the $l^{th}$ hidden layer. 

\section{Experimental Evaluation}

In this section, we first evaluate the online learning performance of the Adaptive TC-LIF neuron model across various sequential modeling benchmarks, including Sequential MNIST (S-MNIST) \cite{le2015simple}, Permutated Sequential MNIST (PS-MNIST) \cite{le2015simple}, and Spiking Heidelberg Digits (SHD) \cite{cramer2020heidelberg}. In the case of the S-MNIST and PS-MNIST datasets, the pixels of images are fed into networks row by row. Subsequently, we conduct comprehensive ablation studies to verify the effectiveness of the proposed modifications to TC-LIF neurons. Finally, the memory efficiency of our proposed online training approach has been compared against the BPTT-based training methodologies.
\subsection{Experimental Setups}
\subsubsection{Hyper-parameter and Network Architecture}
We provide the detailed settings of network structure and hyperparameters of the Adaptive TC-LIF model in Table \ref{Table}. 

\subsubsection{Training Configuration}
We employed the SGD optimizer \cite{robbins1951stochastic} to train the S-MNIST dataset for 300 epochs, with an initial learning rate of 0.08. The learning rate was adjusted using a cosine schedule. For the PS-MNIST and SHD datasets, we utilized the Adam optimizer \cite{kingma2014adam} and trained the networks for 200 epochs. In the case of the PS-MNIST dataset, both the feedforward and recurrent networks had an initial learning rate of 0.0005, which decayed to 0.8 times the previous value after every 15 epochs. Similarly, for the SHD dataset, both networks used an initial learning rate of $5e^{-5}$, which also decayed to 0.8 times the previous value after every 15 epochs. The training process was conducted on an Nvidia GeForce GTX 3090Ti GPU card with 24GB of memory.

\begin{table}[h]
\centering
\caption{Summary of hyperparameters and network architectures.}
\resizebox{0.5\textwidth}{!}{
\begin{tabular}{ccccccc}
\toprule  
\textbf{Dataset} &\textbf{$v_{th}$}& \textbf{$\gamma$} &\textbf{$(a^d, a^s)$} &\textbf{Network}& \textbf{Architecture}& \textbf{Parameters(K)}\\
\midrule  
\multirow{2}{*}{S-MNIST}& 1.0 & 0.5&(0.7,0.8)& feedforward& 64-256-256-10 &85.1\\
                       & 1.0 & 0.5&(0.8,0.9)& recurrent& 64-256-256-10&155.1 \\
\midrule  
\multirow{2}{*}{PS-MNIST} & 1.0 &0.5&(0.7,0.8)& feedforward& 64-256-256-10&85.1 \\
                          & 1.8 &1.0&(0.75,0.85)& recurrent& 64-256-256-10&155.1 \\
\midrule  
\multirow{2}{*}{SHD} & 1.6 &  0.5 &(0.65,0.75)&feedforward& 700-128-128-20&108.8 \\
                     & 1.6 &  0.5 &(0.75,0.85)&recurrent& 700-128-128-20&141.8 \\
\bottomrule 
\end{tabular}}
\label{Table}
\end{table}

\subsection{Superior Sequential Modeling Capability}
We conducted a comparative analysis of the proposed Adaptive TC-LIF model against both LIF and TC-LIF models using three widely-used sequential modeling datasets. The comparison encompassed results obtained through both BPTT (offline) and e-prop (online) training methodologies. As the results presented in Fig. \ref{Fig:superior classification capability}, our proposed Adaptive TC-LIF model consistently outperforms the LIF neuron model in terms of classification accuracy, under the online learning setting. With the same number of parameters and employing the recurrent architecture, the Adaptive TC-LIF model achieved a remarkable classification accuracy of 98.61\%, 98.54\%, and 80.57\% for the S-MNIST, PS-MNIST, and SHD datasets, respectively. These findings highlight the superior sequential modeling capacity of our proposed model.

Furthermore, the experimental results also demonstrate that our online training approach achieves competitive accuracies compared to the offline BPTT algorithm. Specifically, for the S-MNIST dataset, our model when trained online exhibits a minor accuracy gap of only -0.57\% and -0.96\% for the feedforward and feedback architectures, respectively. Similarly, for the PS-MNIST dataset, our online trained feedforward model shows an accuracy gap of -2.23\% compared to BPTT, while the recurrent model even surpasses the accuracy of the vanilla TC-LIF model. The SHD dataset, comprising 250 time steps, poses a significant challenge for online learning algorithms. Notably, the online training results of our Adaptive TC-LIF model remain within 5\% of the BPTT results. These outcomes indicate that our proposed Adaptive TC-LIF neuron model when coupled with the online learning method, strikes a desirable balance between classification accuracy and training efficiency.

\subsection{Ablation Studies}

\subsubsection{Effectiveness of Learnable Decaying Constant}
\begin{figure}[ht]
    \centering
    \includegraphics[width = 1\linewidth]{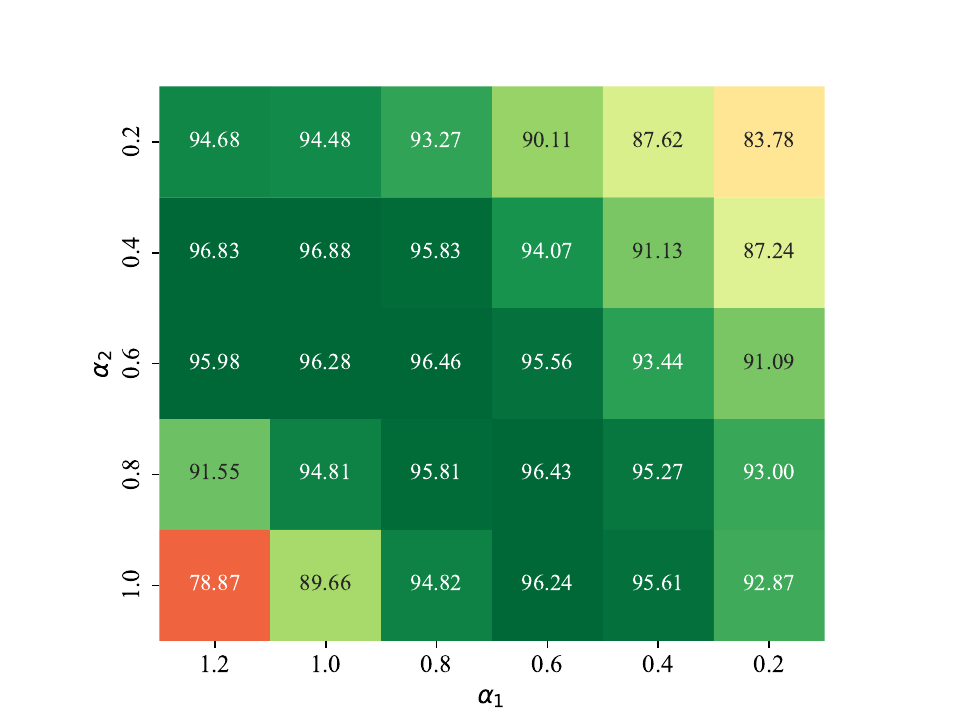}
    \caption{The impact of hyperparameter setting on the test accuracy of the S-MNIST dataset. In this figure, the dark green color indicates high accuracy, while light red indicates low accuracy. The location where both $\alpha_1$ and $\alpha_2$ are equals to 1.0 corresponds to the vanilla TC-LIF model.}
    \label{Figure 2}
\end{figure}

To emphasize the importance of adapting the membrane decaying constants $\alpha_1$ and $\alpha_2$ to suit the online learning scenario, we begin by conducting an ablation study on the vanilla TC-LIF neuron. Specifically, we initialize the hyperparameters $\beta_1$ and $\beta_2$ of TC-LIF neurons to -0.5 and 1, respectively. Furthermore, we conduct a grid search for hyperparameters $\alpha_1, \alpha_2$. The results depicted in Fig. \ref{Figure 2} demonstrate that vanilla TC-LIF neurons, with $\alpha_1 = \alpha_2 = 1$, struggle to adapt effectively to the online learning setting. Conversely, the findings reveal that employing moderate decay constants leads to improved convergence. This enhancement can be attributed to a more balanced contribution between past and new inputs.

\subsubsection{Effectiveness of Time-varying Decaying Constant}

\begin{figure}[ht]
    \centering
    \includegraphics[width = \linewidth]{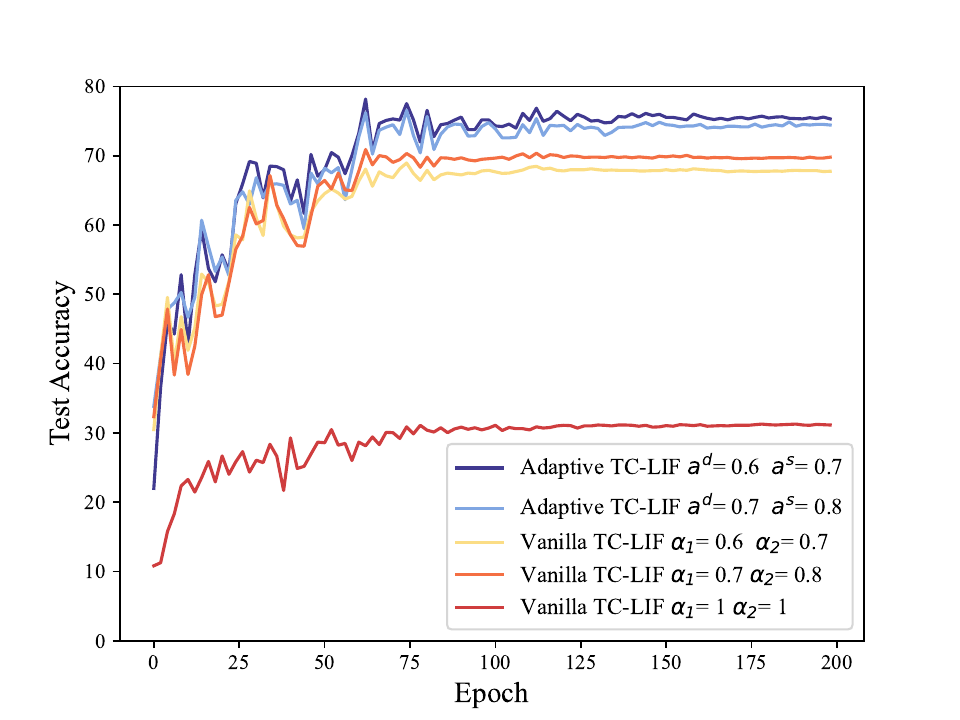}
    \caption{Comparison of the performance with and without time-varying membrane decaying constants on the SHD dataset.}
    \label{Figure}
\end{figure}
To further assess the effectiveness of incorporating time-varying decaying constants into Adaptive TC-LIF neurons, we conducted an ablation study on the SHD dataset. In this experiment, we focused solely on modifying the decaying constant {$A[t]$}, while keeping other hyperparameters constant. As the experimental results depicted in Fig. \ref{Figure}, the proposed Adaptive TC-LIF neuron with time-varying decaying constants exhibited a substantial increase of nearly 10\% in accuracy compared to the best-performing vanilla TC-LIF model. This improvement underscores the crucial role played by time-varying decaying constants in enhancing the learning capabilities of TC-LIF neurons, particularly in the context of long-term sequential modeling tasks.
\subsection{Efficient Memory Consumption}

\begin{figure}[t]
    \raggedleft
    \includegraphics[width = 1\linewidth]{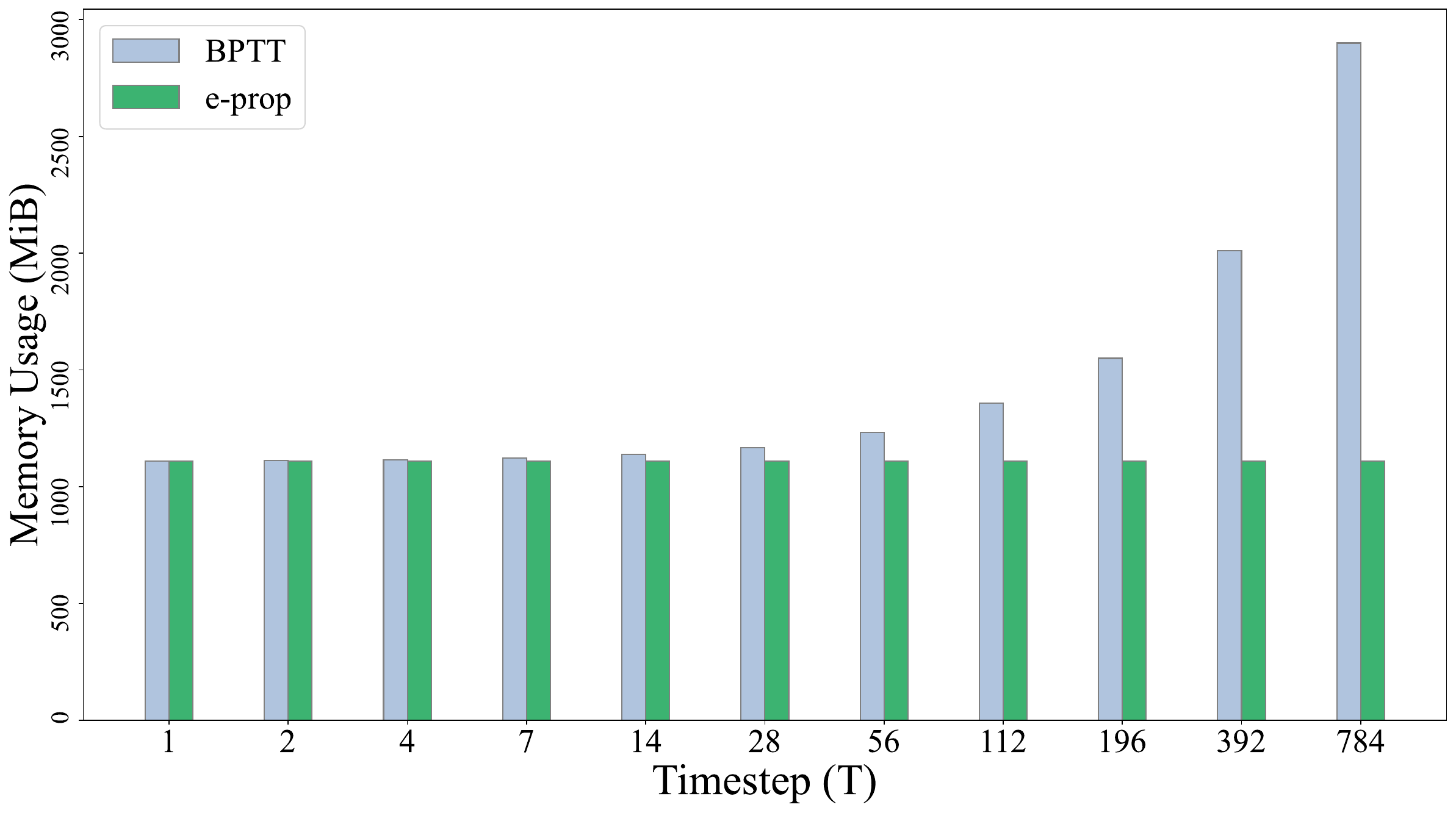}
    \caption{Actual GPU memory usage with varying time sequence lengths on the S-MNIST dataset.}
    \label{Figure memory}
\end{figure}

In theory, the memory cost of BPTT increases linearly with the time window size $T$, denoted as $\mathcal{O} (T)$. This linear relationship poses significant challenges when dealing with long sequence data. In contrast, the e-prop method enables online updates of model parameters, resulting in constant memory consumption. To evaluate the actual memory usage of our proposed Adaptive TC-LIF model during e-prop (online) and BPTT (offline) training, we conducted an empirical study using the S-MNIST dataset. We varied the sequence length from $T = 1$ to $T = 784$. The results, illustrated in Fig. \ref{Figure memory}, validate our theoretical analysis. While BPTT exhibits a linear increase in memory requirements, the e-prop method maintains a relatively stable memory usage regardless of the sequence length. This finding highlights the advantage of our proposed online training approach, especially in scenarios involving extended data sequences.

\section{Conclusion}
In this study, we investigated the feasibility of utilizing an online learning approach to train networks of multi-compartment spiking neurons. Our investigation began with the mathematical derivation of the e-prop algorithm tailored specifically for the TC-LIF neuron model. Additionally, we introduced the Adaptive TC-LIF neuron model by carefully analyzing the limitations of the original hyperparameter space used in the vanilla TC-LIF neuron for online training scenarios. Our approaches showcased comparable performance to the vanilla TC-LIF model achieved using the offline BPTT method, substantially surpassing the performance of LIF models trained with the e-prop algorithm. Furthermore, our proposed online training approach requires consistent memory usage, regardless of the sequence length. This highlights the superior memory efficiency of our approach. As a result, our study has laid a solid foundation towards enabling the training of high-performance multi-compartmental SNNs on emerging neuromorphic hardware.

\bibliographystyle{IEEEtran}
\bibliography{references.bib}

\end{document}